\documentstyle[12pt]{article}
\title{Compositionality for Presuppositions over Tableaux\footnote{
Presented in:
{\em Computational Logic for Natural Language Processing},
(A Joint COMPULOG/ELSNET/EAGLES Workshop),
April 3-5, 1995, Edinburgh.}}
\author{Pablo Gerv\'{a}s \\
{\small Department of Computing}\\
{\small Imperial College of Science, Technology and Medicine}\\
{\small  London SW7 2BZ, UK.}\\
{\small e--mail: pg2@doc.ic.ac.uk}
}

\begin{document}
\maketitle

\begin{abstract}

Tableaux originate as a decision method for a logical language.
They can also be extended to obtain a structure that
spells out all the information in a set of sentences in terms of
truth value assignments to atomic formulas that appear in them.
This approach is pursued here.  Over such a structure,
compositional rules are provided for obtaining the
presuppositions of a logical statement from its atomic subformulas
and their  presuppositions.
The rules are based on classical logic semantics
and they are shown to model the  behaviour of presuppositions
observed in natural language sentences built with
{\em if \ldots then}, {\em and} and {\em or}.
The advantages of this method  over existing
frameworks for presuppositions are discussed.

\end{abstract}

\vspace{.5cm}


\begin{section}{The Problem of Compositionality for Presuppositions}

This work is concerned with the problems of adding the concept
of presupposition to a logical language.
Although presupposition originates as a natural language
phenomenon, for the purposes of the present work
sentences will be represented as propositions of a
logical language.
At this level of granularity, presupposition can simply be represented
as a relation between sentences.
For instance, the sentence
{\em 	The typewriter is working}
can be said to presuppose:
{\em 	There is a typewriter.}

Presupposition has the additional property (often used to characterize it)
that the negation of a sentence has the same presuppositions as
the sentence itself.
For example, the sentence
{\em 	The typewriter is not working}
has the same presupposition as its positive counterpart given above
(namely, that there is a typewriter).
The reader is referred to the literature on presupposition for a
wider analysis of the implications of this property
\footnote{The most important consequence is that attempts to model
presupposition as an entailment lead to  characterizations
where all presuppositions are tautologies in the classical sense,
which is an undesirable result.  This is based on the argument that
if $ \alpha \vdash \beta $ and $ \neg \alpha \vdash \beta $
then it must be the case that $  \vdash \beta $.  Because
presuppositions are observed to be informative in their actual
use by language speakers, this way of modelling them is
not useful.}.

A formal language has the property of {\em compositionality}
if it is possible to describe the meaning of a complex  expression
of the language in terms of the meaning of its parts.
It is considered a desired property for any formal language.
When a logical
statement is composed from propositions that presuppose other
propositions, it should be possible to  describe the presuppositions of the
resulting complex expression in terms of its parts and the presuppositions
of its parts.
If one takes the natural language connectives {\em if \ldots then },
{\em and  },  and {\em or  } to be related to material conditional,
conjunction, and disjunction, natural language examples
provide some clues as to what the behaviour of presupposition
should be.
Sentences (1), (2) and (3)
presuppose
{\em There is a typewriter}; while sentences (4), (5)  and (6) do not.

\begin{quote}

(1) {\em  If the typewriter is blue then Sue will be happy.}

(2) {\em  If  you are in Sue's office then the typewriter is blue.}

(3) {\em  Either the typewriter is blue or the chair is.}

(4) {\em  If there is a typewriter then the typewriter is blue.}

(5) {\em  Either there isn't a typewriter or the typewriter is blue.}

(6) {\em  Either the typewriter is blue or there isn't a typewriter.}

\end{quote}

The behaviour of presuppositions of sentences of this form has
traditionally been studied as part of the
{\em projection problem for presuppositions}, which is concerned with
describing the presuppositions of a sentence in terms of the
presuppositions of its subordinate clauses.  The constructions
considered in the projection problem involve nested subordination
(verbs of propositional attitude, factive verbs) beyond
the natural language connectives treated here.

\end{section}

\begin{section}{Previous Work}

The framework presented in this work is compared with
three other approaches.

Beaver \cite{db:kop,db:wfd} presents a solution based on update semantics.
This account  originates from an approach to
the projection problem that concentrates on how the
presuppositions of a
compound sentence are obtained
from the presuppositions of its components
(Karttunen \cite{kt:pcs,kt:plc},
Karttunen and Peters \cite{kp:cim} and Heim \cite{ih:sdi,ih:ppp}).
The projection problem is rephrased in terms
of what it takes for a context (represented as a set of possible worlds)
to satisfy a presupposition.  This question is answered by
defining  for each connective  updates of compounds in terms of
sequences of simpler updates
that involve only  their components.
However, there are several problems: no satisfactory account for
the behaviour of disjunction is provided;
the  defeasibility of presuppositions is not addressed;
and presupposition as an informative operation is only
allowed as a repairing  modification to the original framework.

Mercer \cite{rm:dlp,rm:pdl} applies a default logic approach.
This account  originates from an approach to
the projection problem that concentrates
on how presupposition as an inference
is defeated when inconsistent with more firmly established information
(Gazdar \cite{gg:ppc,gg:spp}).
Presuppositions (of negative sentences) are represented as normal
defaults.
Mercer's method follows Gazdar in taking into account the internal
structure of sentences only  in terms of Gricean pragmatic implicatures
of the sentence.
In order to avoid excessive production of presuppositions,
Mercer needs to take into account
that an assertion of $ \alpha \vee \beta $ or
$ \alpha \rightarrow \beta $
carries with it the assumption that the speaker does not know
$ \alpha $,$  \neg \alpha $, $ \beta $, or $ \neg \beta $,
that is, that he considers an information state where
all of them are open possibilities.
The consistency checks required for defaults must consider these implicatures.
The method obscures the issue
of how the compositionality for truth values and the
compositionality for presupposition are related.

There is an alternative approach (van der Sandt \cite{vs:par})
within the framework of DRT.
This approach considers presupposition as anaphora.
In this approach, interpretation of presupposition is
reformulated as a search for an anaphoric referent.
This solution is essentially linguistic.
It captures an essential
aspect of the nature of presupposition (its anaphoric nature),
but it fails to address
its direct relation with the semantics required for interpreting these
connectives in a logical sense.

\end{section}

\begin{section}{The Tableau Interpretation}

\begin{subsection}{The Basic Framework}

Where language is simplified to a set of sentences so that
the internal construction of each sentence does not play a role,
the representation of presupposition can be  restricted to defining
the ordered pairs of sentences for which this relationship holds.
I assume that such a  relation of presupposing is given for the
atomic formulas of the language.
To make this information conspicuous without introducing
too many definitions, I notate the fact that
`$ \alpha $ presupposes $ \beta $ ' by writing each instance
of $ \alpha $ as $ \alpha^{\beta} $.

In terms of this notation, the behaviour of presupposition
concerning negation has the following implication:
	$ \neg (\alpha^{\beta}) \equiv (\neg \alpha)^{\beta} $
(For simplicity, I leave out the parentheses in these cases from now on.)

For complex sentences the relation of presupposing  has to be worked out,
ideally in a compositional way.
This is achieved in the next section by defining rules that
govern the compositionality.
In order to obtain a simpler formulation of these rules,
a specific representation of the logical structure of the
connectives is required.
This representation is described in the present section.

The logical connectives I am considering have their own
definition of compositionality with respect to truth value.
In the present framework, these definitions are represented
as tableaux expansion rules.

Tableau expansion rules:

\footnotesize

\[
\begin{array}{lcccccc}
	\neg-rules)
& &
	\begin{array}{c}
		\underline
		{\neg \neg \phi} \\
		\phi
	\end{array}
& &

& &
\\
\\
	\alpha-rules)
& &
	\begin{array}{c}
		\underline
		{\alpha_{1} \wedge \alpha_{2} } \\
		\alpha_{1}  \\
		\alpha_{2}
	\end{array}
& &
	\begin{array}{c}
		\underline
		{ \neg (\alpha_{1} \rightarrow \alpha_{2} ) } \\
		\alpha_{1}  \\
		\neg \alpha_{2}
	\end{array}
& &
	\begin{array}{c}
		\underline
		{ \neg (\alpha_{1} \vee \alpha_{2} ) } \\
		\neg \alpha_{1}  \\
		\neg \alpha_{2}
	\end{array}
\\
\\
	\beta-rules)
& &
\begin{array}{c}
	\beta_{1} \vee \beta_{2} \\
	\overline{
	\begin{array}{ccc}
		\beta_{1} 	& \neg \beta_{1}	& \beta_{1} \\
		\beta_{2}	& \beta	_{2}		& \neg \beta_{2}
	\end{array}
	}
\end{array}
& &
\begin{array}{c}
	\beta_{1} \rightarrow \beta_{2} \\
	\overline{
	\begin{array}{ccc}
		\neg \beta_{1}	& \neg \beta_{1}	& \beta_{1} \\
		\beta_{2}	& \neg \beta_{2}	& \beta_{2}
	\end{array}
	}
\end{array}
& &
\begin{array}{c}
	\neg (\beta_{1} \wedge \beta_{2}) \\
	\overline{
	\begin{array}{ccc}
		\neg \beta_{1} 	& \neg \beta_{1}	& \beta_{1} \\
		\neg \beta_{2}	& \beta	_{2}		& \neg \beta_{2}
	\end{array}
	}
\end{array}
\end{array}
\]

\normalsize

Classical negation operates over
the language wherever reasoning about negated sentences is
required.

Some additional definitions are needed to introduce this
way of understanding the representation.
Apart from the expansion  rules, these definitions follow
existing tableau frameworks for propositional logic.
They include definitions of:
a tableau,
a branch of a tableau,
branch closure, and
closure for a tableau.
(For standard definitions, see Fitting \cite{mf:fol}).

\end{subsection}

\begin{subsection}{Coverage Property}

The definition of tableau expansion rules ensures that
these tableau obey a special property.

\begin{quote}

{\em Coverage Property:}

If one branch of a tableau holds
a sentence $ \delta $, then every (open) branch of that
tableau will hold either $ \delta $ or $ \neg \delta $.

\end{quote}

This property can be seen to hold:
1) it applies to each one of the expansion rules,
2) the procedure for adding sentences to a tableau is
defined in terms of adding the new sentence to every (open)
branch (and then applying the expansion rules to it).

A consequence of this property in terms of the semantics,
is that each branch of the tableau contains a complete atomic
truth-value assignment that makes the sentence true.
The structures that result are equivalent to classical truth tables.
A tableau formulation is retained in spite of this fact because it
takes into account that lines of the truth table that become inconsistent
as more information is added are dropped out of the reckoning
(as an effect of branch closure).

\end{subsection}

\begin{subsection}{The Formalism as a Decision Method for the Logic}

The present framework differs from traditional tableaux only
in the definition of $ \beta $ expansion rules.
The traditional definition of $ \beta $ expansion rules is not suitable
for presupposition because it specifies the alternatives in
the minimal form that preserves soundness and completeness
of the logical calculus.   This is done in order to simplify
the computation of logical consequences.
The information that is not specified explicitly as a result of
this policy
does not affect logical consequences, but it is relevant
to presupposition behaviour as described by the compositionality
rules given here.

The tableaux given here (presuppositional tableaux or PT)
can be used as a decision method
in the same way as traditional tableaux
(semantic tableaux or ST).
A simple way of showing this may be to
show that PT tableau are equivalent to ST tableaux.
Given that the definitions of tableau, tableau for a set of sentences,
closure, proof and refutation are the same in the PT
framework as in ST tableaux, the discussion concerns
only the expansion rules.  Of these, only $ \beta $-rules
differ from one framework to the other.
It is enough to show that under the circumstances
under which ST tableau for a $ \beta $-formula becomes closed,
the PT tableau for the same $ \beta $-formula also becomes closed
(and viceversa).

\end{subsection}

\end{section}

\begin{section}{Compositionality of Presupposition over Logical Connectives}

\begin{subsection}{Compositionality Rules for Presupposition}

Over the representation of the logical language given in
the previous section,
compositionality of presupposition can be defined.

Two issues need to be dealt with:
1) how the presuppositions of a branch
are ascertained
from the presuppositions of the atomic propositions in it, and
2) how the presuppositions of branches of  the same tableau
interact.

I notate a branch as $ \Delta $.
I use $ \beta \in \Delta $ as shorthand for `the proposition $ \beta $
appears in the branch $ \Delta $.  The notation for presupposition
is extended to branches so that $ \Delta^{\beta} $ stands
for `the branch $ \Delta $ presupposes $ \beta $'.

The compositionality rules for presupposition do not take into account
presuppositional information
from branches of a tableau that are closed.

\begin{quote}

{\bf Rule 1}

For an open branch $ \Delta $ such that $ \alpha^{\beta} \in \Delta $,
$ \Delta^{\beta} $ unless
i) $ \neg \beta  \in \Delta $, or
ii) $ \beta \in \Delta $, or
iii)  there is some $ \delta^{\neg \beta} \in \Delta $.

{\bf Rule $ 2 $}

For a tableau $ \Gamma $ with at least one
open branch $ \Delta $,  $ \Gamma^{\beta} $ iff $ \Delta^{\beta} $.

\end{quote}

The determination of presuppositions of a branch
is handled in a simple way by considering
that presupposition is weaker than assertion, so that it survives
only in cases where it does not overlap or clash with any other information.
Conditions 1.i and 1.ii  capture the intuitive observation
that presupposition is only informative when it concerns
a proposition that does not appear in any  literal already
in the branch.
Condition 1.iii allows the presuppositions of a branch to survive
only when there is no conflict between them.

The interaction between presuppositions of different branches
can be shown to be trivial in the present
framework by application of the Coverage Property.
By the Coverage Property, different branches wil have the same
atomic propositions and they will differ only in that
each atomic proposition may appear negated in one branch and
appear unnegated in another.
In terms of Rule 1, these differences would affect only
$ \alpha $ and $ \delta $ as possible origins of presuppositions
and $ \beta $ in conditions i) and ii).
Because the relation of presupposing is given for atomic propositions,
if $ \delta^{\neg \beta} $ in  one branch, it is not possible to have
$ \delta^{\beta} $ in a different branch.
The results of applying rule 1 are the same when any of
these propositions is exchanged for its negation\footnote{
For $ \alpha $ and $ \delta $,
the negated and the unnegated version have the same presupppositions;
for $ \beta $, substituting with the negation is equivalent to interchanging
conditions i) and ii). }.
This can be used to show that
if a tableau has a branch that presupposes a proposition,
then every branch of the tableau will presuppose that proposition.
Conversely, if a tableau has a (open)  branch that does not
presuppose a proposition, then no branch of the tableau will
presuppose that proposition.

\end{subsection}

\begin{subsection}{Simple Sentences}

I give the first few examples in detail
to show how the rules operate.

Example (1) is a case of behaviour of presuppositions originating
in the antecedent of a conditional.
Sentence
(1) {\em  If the typewriter is blue then Sue will be happy}
corresponds to the following representation.

\footnotesize

\[
\begin{array}{c}
	\alpha^{\beta} \rightarrow \gamma	\\
	\overbrace{
	\begin{array}{ccc}
		\neg \alpha^{\beta} & \neg \alpha^{\beta} & \alpha^{\beta} \\
		\gamma		& \neg \gamma		& \gamma
	\end{array}
	}
\end{array}
\]

\normalsize

By application of rule 1, every one of the alternatives
presupposes $ \beta $.  None of  the conditions holds for any branch.
By application of rule 2 to any one of them,  $ \beta $ is obtained
as a presupposition of the compound.   This matches expected behaviour.

The sentence
(7) {\em  Bill regrets that there is no hot water left}
presupposes
(8) {\em There is no hot water left.}
The problem is to determine what the presuppositions
are for sentence (9) {\em  If Mary has had a bath, then Bill regrets
that there is no hot water left.}
Assume sentence (9) has the form $ \alpha \rightarrow \delta^{\beta} $.
The representation for this sentence in this framework would be:

\footnotesize

\[
\begin{array}{c}
\alpha \rightarrow \delta^{\beta}	\\
\overbrace{
	\begin{array}{ccc}
	\neg \alpha	& \neg \alpha	& \alpha \\
	\delta^{\beta}	& \neg \delta^{\beta}	& \delta^{\beta}
	\end{array}
	}
\end{array}
\]

\normalsize

Application of the rules predicts a presupposition
$ \beta $ for the sentence.

Example (4) is a case of behaviour of presuppositions originating
in the consequent of a conditional, in the particular case where
the presupposition itself forms the antecedent.  Sentence
(4) {\em  If there is a typewriter then the typewriter is blue}
corresponds to the following representation.

\footnotesize

\[
\begin{array}{c}
	\beta \rightarrow \alpha^{\beta}	\\
	\overbrace{
	\begin{array}{ccc}
		\neg \beta	& \neg \beta		& \beta	\\
		\alpha^{\beta}	& \neg \alpha^{\beta}	& \alpha^{\beta}
	\end{array}
	}
\end{array}
\]

\normalsize

By rule 1, none of the alternatives under the compound presuppose
$ \beta $
(condition 1.i holds for the first two columns,
condition 1.ii holds for the third one).
This agrees with the intuitive behaviour that
had to be modelled.

Disjunction  presented the most problems in update semantics
attempts to model intuitive behaviour in terms of compositionality.
Example (5) {\em  Either there isn't a typewriter or the typewriter is blue}
shows the case where the presuppositions of one
of the disjuncts are not presuppositions of the disjunction.
This sentence corresponds to a structure that is equivalent to
that for the conditional of
example (4), and gives no presupposition for the compound.  This matches
the required behaviour.

Unlike update semantics methods, this approach can also handle
the symmetrical version of the disjunct with no problems.
For the sentence in example
(6) {\em  Either the typewriter is blue or there isn't a typewriter}
the corresponding expansion contains the same literals in each branch
but in a different order.  Since order in the branch plays no role
in the compositionality rules\footnote{
Ordering does seem to play a role in the intuitive
validity of conjunctions
({\em There is a typewriter and the typewriter is blue} is acceptable,
{\em The typewriter is blue and there is a typewriter} is harder to accept).
This issue is discussed further in \cite{pg:lcp}.},
the predictions are the same as for the previous case.

The method can also handle the case of disjunctions with
contradictory presuppositions.
Take for instance the sentence in example (10)
 {\em Either Bill has started smoking or Bill has stopped smoking.}
This has the following representation:

\footnotesize

\[
\begin{array}{c}
	\delta^{\neg \beta} \vee \alpha^{\beta}	\\
	\overbrace{
	\begin{array}{ccc}
	\delta^{\neg \beta} & \delta^{\neg \beta} & \neg \delta^{\neg \beta}\\
	\alpha^{\beta} & \neg \alpha^{\beta} & \alpha^{\beta}
	\end{array}
	}
\end{array}
\]

\normalsize

In this case, the method  predicts
(condition 1.iii)  no presupposition
for any column.
Application of rule 2 gives no presupposition for the
whole compound.

\end{subsection}

\begin{subsection}{Complex Sentences}

Complex sentences are treated as follows.
First the sentence is expanded into a tableaux by
application of the expansion rules.
Then the compositionality rules are
applied to the resulting tableau  in order to obtain the presuppositions
of the sentence.

Sentence (11) {\em If John is married and he has children, then his
children are at school  }
can act as an example.
Assuming a logical form for this sentence
$ (\alpha \wedge \beta ) \rightarrow \delta^{\beta}  $,
the tableau for this sentence would be:

\footnotesize

\[
\begin{array}{c}
(\alpha \wedge \beta ) \rightarrow \delta^{\beta} 	\\
\overbrace{
\begin{array}{ccc}
\neg(\alpha \wedge \beta) & \neg(\alpha \wedge \beta) & \alpha \wedge \beta\\
\delta^{\beta} & \neg \delta^{\beta}      & \delta^{\beta} \\
\overbrace{
\begin{array}{ccc}
\neg \alpha	& \neg \alpha 	& \alpha  \\
\neg \beta	& \beta 	& \neg \beta
\end{array}
}
&
\overbrace{
\begin{array}{ccc}
\neg \alpha	& \neg \alpha 	& \alpha  \\
\neg \beta	& \beta 	& \neg \beta
\end{array}
}
&
\overbrace{
\begin{array}{c}
\alpha  \\
\beta
\end{array}
}
\end{array}
}
\end{array}
\]

\normalsize

The presupposition rules applied to this tableau predict
no presupposition. (Rule 1 fails for all branches).

\end{subsection}

\begin{subsection}{Discourses}

When logical statements of the type considered so far are
strung into a sequence of assertions, a {\em discourse }
is obtained.
An appropriate treatment of discourses is required to study
the effect of context on presuppposition interpretation.
A discourse is represented by the tableaux for the set
of formulas in it.

A problematic case studied by Beaver \cite{db:wfd}
  is based on example  (9) above.
Suppose it is understood in the context  that
whenever Mary has a bath, she uses up all the hot water.
This can be represented by assuming that the context
already holds sentence (12)
 {\em  If Mary has had a bath, then there is no hot water left},
with a logical form $ \alpha \rightarrow \beta $.
A representation for the discourse  (12),(9) or
$ \alpha \rightarrow \beta,\alpha \rightarrow \delta^{\beta} $
would be the following tableau:

\footnotesize

\[
\begin{array}{c}
\mbox{\boldmath $ \alpha \rightarrow \beta $}\\
\overbrace{
\begin{array}{ccc}
\neg \alpha	& \neg \alpha	& \alpha \\
\beta 		& \neg \beta 	& \beta \\
	\mbox{\boldmath $ \alpha \rightarrow \delta^{\beta} $} &
	\mbox{\boldmath $ \alpha \rightarrow \delta^{\beta} $} &
	\mbox{\boldmath $ \alpha \rightarrow \delta^{\beta} $} \\
	\overbrace{
	\begin{array}{ccc}
	\neg \alpha	& \neg \alpha	& \alpha \\
	\delta^{\beta}	& \neg \delta^{\beta}	& \delta^{\beta}
	\end{array}
	}
	&
	\overbrace{
	\begin{array}{ccc}
	\neg \alpha	& \neg \alpha	& \alpha \\
	\delta^{\beta}	& \neg \delta^{\beta}	& \delta^{\beta}
	\end{array}
	}
	&
	\overbrace{
	\begin{array}{ccc}
	\neg \alpha	& \neg \alpha	& \alpha \\
	\delta^{\beta}	& \neg \delta^{\beta}	& \delta^{\beta}
	\end{array}
	}
\end{array}
}
\end{array}
\]

\normalsize

Over this representation it can be seen that the rules correctly
predict no presupposition for the discourse as a whole,
even though sentence (9) on its own
did have a presupposition.

\end{subsection}

\begin{subsection}{Traditional Presuppositional Concepts in
Terms of Tableaux}

The tableaux framework allows definition of
some of the traditional concepts that surround presupposition.

The presupposition  $ \phi $ of a presuppositional sentence
$ \delta^{\phi} $ added to a tableau $ \Gamma $
is {\em satisfied} if the tableau $ \Gamma \cup \{ \neg \phi \} $
is closed.

The presupposition  $ \phi $ of a presuppositional sentence
$ \delta^{\phi} $ added to a tableau $ \Gamma $
is {\em canceled} if the tableau $ \Gamma \cup \{ \phi \} $
is closed.

These two definitions correspond to the intuitive concepts of
satisfaction and cancelation.
However, it is clear that there will be cases when
some branches of a tableau are closed by $ \phi $ and
some by $ \neg \phi $.  These hybrid cases
between satisfaction and cancelation escape the simpler analysis
and give rise to the need for projection rules.
In their simplest manifestation,  hybrid cases occur
as the traditional  cases of
problematic projection.
These involve sentences (4), (5) and  (6) given above.
More complex manifestations concern discourses where the effect
of context plays a role in the interpretation of presupposition.
The discourse constructed with sentences (12) and  (9)
is an instance of these cases.
In all these examples  it holds that for any of the tableau representations
some branches of the tableau are
closed by the presupposition involved and
some by its negation.
Under those circumstances, the traditional definitions
of satisfaction and/or cancelation could not account for the
resulting presuppositional behaviour.

The present framework achieves this by allowing presupposition
to be blocked locally by either $ \beta $ or $ \neg \beta $
(conditions 1.i and 1.ii ).

\end{subsection}

\end{section}

\begin{section}{Critical Analyisis}

\begin{subsection}{Advantages over van der Sandt}

The framework presented here is closely related to
that proposed by van der Sandt.
Both can be interpreted as  a  branching
structure for a sentence based on the connective words that appear
in it.
The advantages of this framework are that
1) it makes explicit the semantics that are
attributed to the connectives, and
2) logical consistency (both at sentence level and at branch level)
and logical consequence are explicitly taken into account in the framework.

\end{subsection}

\begin{subsection}{Advantages over Beaver}

The behaviour of all the connectives (including disjunction)
is described satisfactorily by the rules given.
No specific rules for each connective are required.
The compositionality rules rely on the semantics in
general terms.  As a result, the same compositionality
rules may be applied to other connectives if
their semantics can be represented in the same framework
in a way that preserves the Coverage Property.

The general  approach that starts with Karttunen and
leads to Beaver is criticized for not being able to justify their
choice of method for obtaining presuppositions of compounds on grounds other
than that it describes the behaviour.
In the present framework the method is given simply by
attributing a weak informative  status to presupposition (so that
it is overridden whenever it overlaps or conflicts with
explicit  information or other presuppositions).

\end{subsection}

\begin{subsection}{Advantages  over Mercer}

In the present framework the information that Mercer
must introduce as pragmatic implicatures
can be read off
the representation.
An expansion rule like:

\[
\begin{array}{c}
	\beta_{1} \vee \beta_{2} \\
	\overline{
	\begin{array}{ccc}
		\beta_{1} 	& \neg \beta_{1}	& \beta_{1} \\
		\beta_{2}	& \beta	_{2}		& \neg \beta_{2}
	\end{array}
	}
\end{array}
\]

captures the idea that each of $ \beta_{1} $, $ \neg \beta_{1} $, $ \beta_{2} $
and $ \neg \beta_{2} $ is  valid in at least one of the possible
alternatives given.
This property of the present framework is a direct result of the
insistence on taking the semantics -- as given by the
internal structure of the sentence in terms of connectives --
into account.

The method followed by the compositional rules
is quite close to Gazdar's method for
computing presuppositions simply in terms of consistency with the
context.  However, it presents two major innovations:
1) it allows both satisfaction and cancelation (presuppositions
disappear when inconsistent and/or when already present), and
2) it applies the procedure locally.
The combination of these two innovations allows adequate
treatment of hybrid cases.

\end{subsection}

\end{section}

\begin{section}{Conclusions}

Presuppositional behaviour in sentences can be described in relation to the
implicit logical structure represented by the appearance of
natural language connectives {\em if \ldots then },
{\em and  },  and {\em or  } in the sentences to be interpreted.

The framework presented here provides a
method for determining the behaviour of presuppositions
of complex logical statements in a compositional manner.

The predictions of the proposed method match the observed behaviour
in traditional examples.

The semantics used in the present framework are chosen to ensure
that all the different valid alternatives implied by a sentence are
listed explicitly in a tableau for that sentence,
and all atomic formulas involved appear (either negated or not)
in every branch (Coverage Property).
These constraints on the semantics allow all the predictions of
the framework to be explained in terms of two basic assumptions:
presuppositions are only considered informative where they
do not overlap with  or contradict asserted information,
and where they do not conflict with other presuppositions.

There are some problems left unsolved.
The present framework covers both information explicitly
represented in the logic and presuppositional information,
but does not allow a unified treatment;
and it  does not address the defeasible
nature of certain presuppositions.
In Gerv\'{a}s \cite{pg:lcp}, these problems are addressed by
providing tableau expansion rules for presupposition.
This makes presuppositional information explicit in
the framework.
Because the issue of defeasibility
is closely related to consistency checking, the  framework
proposed here presents advantages for this purpose
by  having the semantics
of each connective made fully explicit.

\end{section}

\end{document}